\documentclass[10pt,journal,compsoc]{IEEEtran}
\ifCLASSOPTIONcompsoc
  \usepackage[nocompress]{cite}
\else
  \usepackage{cite}
\fi
\ifCLASSINFOpdf
\else
\fi
\usepackage{graphicx}                
\usepackage{booktabs}                  
\usepackage{array,multirow}

\usepackage{amsmath}
\usepackage{url}

\begin{document}
%
\title{Real Time Egocentric Segmentation for Video-self Avatar in Mixed Reality}

\author{Ester~Gonzalez-Sosa,
        Andrija~Gajic,
        Diego~Gonzalez-Morin,
        Guillermo~Robledo, \\
        Pablo Perez
        and~Alvaro~Villegas,
\IEEEcompsocitemizethanks{\IEEEcompsocthanksitem E. Gonzalez-Sosa, P. Perez, R. Kachach, and A. Villegas are Nokia Bell Labs Spain,
Maria Tubau 9, Madrid, 28050. 
E-mail: ester.gonzalez@nokia-bell-labs.com. 
\IEEEcompsocthanksitem Andrija Gajic and Guillermo Robledo made their contributions during an internship at Nokia}
\thanks{}}
\markboth{}%
{Shell \MakeLowercase{\textit{et al.}}: Bare Demo of IEEEtran.cls for Computer Society Journals}
\IEEEtitleabstractindextext{
\begin{abstract}\rightskip=0pt In this work we present our real-time egocentric body segmentation algorithm. Our algorithm achieves a frame rate of 66 fps for an input resolution of 640x480, thanks to our shallow network inspired in Thundernet's architecture. Besides, we put a strong emphasis on the variability of the training data. More concretely, we describe the creation process of our Egocentric Bodies (EgoBodies) dataset, composed of almost 10,000 images from three datasets, created both from synthetic methods and real capturing. We conduct experiments to understand the contribution of the individual datasets; compare Thundernet model trained with EgoBodies with simpler and more complex previous approaches and discuss their corresponding performance in a real-life setup in terms of segmentation quality and inference times. The described trained semantic segmentation algorithm is already integrated in an end-to-end system for Mixed Reality (MR), making it possible for users to see his/her own body while being immersed in a MR scene.
\end{abstract}
}

\maketitle

\IEEEdisplaynontitleabstractindextext

%
\IEEEpeerreviewmaketitle

\IEEEraisesectionheading{\section{Introduction}\label{sec:introduction}}

%
%
%
%

\IEEEPARstart{I}{t} has been already a decade since the emergence of deep neural networks \cite{krizhevsky2012imagenet}. Considered as a breakthrough in the general field of machine learning, they have revolutionized different research areas \cite{lecun2015deep} such as natural language processing, speaker recognition, recommendation systems, or computer vision. Concerning the latter, there are many related tasks whose state-of-the-art solutions are based on convolutional neural networks (CNN), e.g. image classification \cite{khan2021transformers}, object detection \cite{girshick2015fast}, or \emph{semantic segmentation}, among others. Semantic segmentation is the task of, given an input image, assigning class information at pixel-wise level. Unlike image classification, or object detection, where groundtruth information is simply a text label, semantic segmentation requires the groundtruth information for every pixel with an extremely high labelling associated cost.

\begin{figure}[t]
  \centering
  \includegraphics[width=1.0\linewidth]{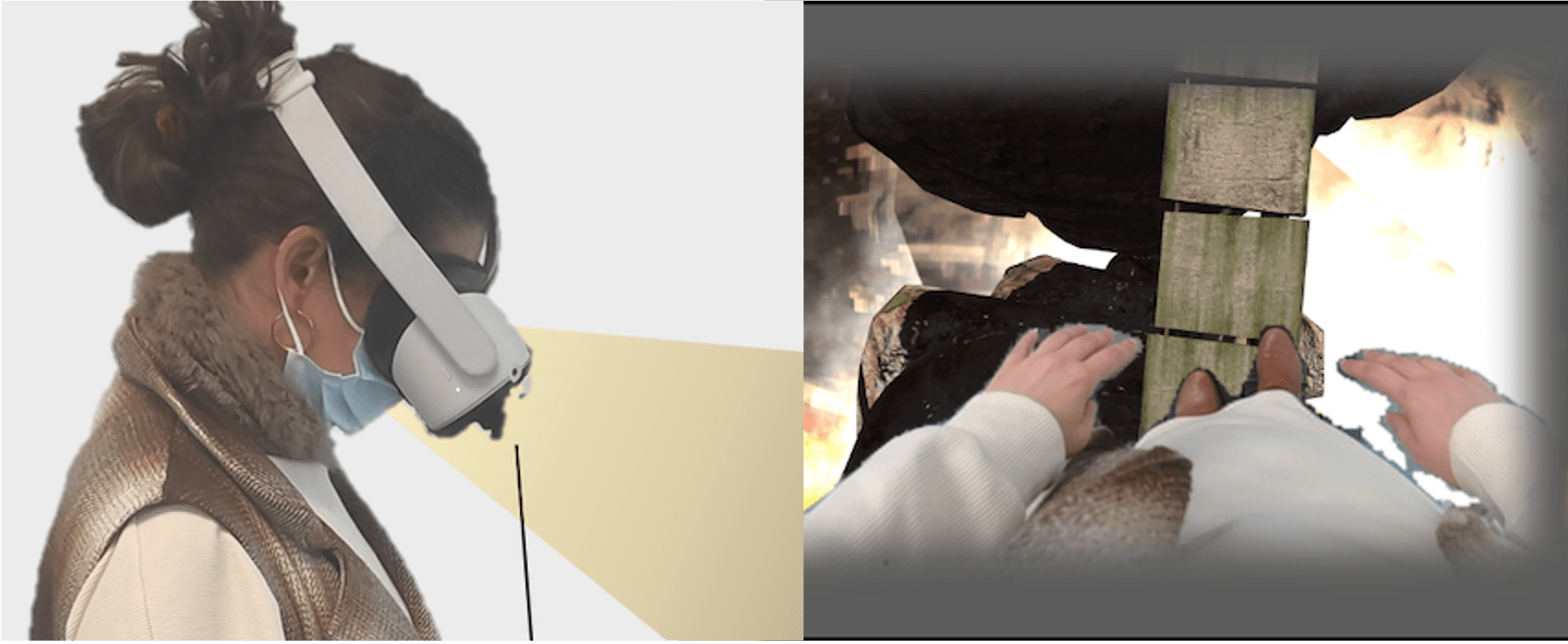}
  \caption{Performing semantic segmentation in real time is crucial for Mixed Reality applications. Left) user wearing VR googles with a stereo RGB camera attached in front of it; right) view of the user when wearing the goggles.}
	\label{fig:mr}
\end{figure}

In the last few years, Mixed Reality (MR) has benefited from the use of semantic segmentation. In MR experiences, users usually see themselves in the form of a virtual graphical avatar. One alternative approach would be to use video-based self-avatars, by segmenting body limbs (arms, legs, or whole body) from the egocentric vision captured from a camera attached to a Virtual Reality (VR) device, as in Fig ~\ref{fig:mr}. Previous approaches for bringing real bodies into MR have been based on: $i)$ color information, allowing users to see their own hands/ bare arms \cite{villegas2020realistic}; $ii)$ depth \cite{rauter2019augmenting}, by segmenting anything below a certain distance threshold or even deep learning to segment bare/clothes arms \cite{gonzalez2020enhanced}, or whole bodies \cite{pigny2019using}. However, those recent methods based on deep learning still fail at reaching sufficient execution speed.

The success of a solution for segmenting egocentric whole bodies using deep learning relies on two critical requirements: $i)$ real time segmentation and $ii)$ high segmentation quality. At least $60$ fps are required for a semantic segmentation algorithm to be seamlessly integrated in a MR application\footnote{https://developer.oculus.com/resources/oculus-device-specs/}. Additionally, the segmentation quality also needs to be good enough to visualize the user's body parts accurately with none or few false positives. In this use-case, high quality segmentation needs to be achieved not only in benchmarking datasets, but even more important, in realistic setups. In general, high segmentation quality requires a heterogeneous training data reflecting real-world settings\footnote{https://www.forbes.com/sites/gilpress/2021/06/16/andrew-ng-launches-a-campaign-for-data-centric-ai/?sh=51a540ea74f5}. Another additional drawback to bear in mind is the extremely high labelling cost associated to segmentation groundtruth, since class information is given at pixel level.

In this work we contributed with a highly shallow architecture that meets the real time requirements while achieving to accurately segment the user's own body in many diverse scenarios, with different illumination conditions, scenes, user demographics, etc. This high quality segmentation is achieved by putting a strong emphasis on the variability of the training data. More in particular, we have used almost $10,000$ images, coming from three different data sources: $i)$ EgoHuman: a semi synthetic dataset inspired by our previous work \cite{gonzalez2020enhanced},  where we contribute with a more sophisticated method for seamlessly combining foreground and background and additional images containing lower limb parts, $ii)$ a subset of THU-READ dataset \cite{tang2018multi}, originally created for action recognition and whose segmentation groundtruth was generated in a previous work \cite{gonzalez2022real}, and $iii)$ EgoOffices: an egocentric dataset captured in many different real-life settings with more than $25$ people, actions, objects and scenarios, for which we have developed an egocentric capture kit to obtain a very extensive dataset. The resulting combined dataset is referred hereinafter as the EgoBody dataset: an egocentric semantic segmentation dataset with a wide range of variability in terms of users, skin color, illumination, scenes, etc. The corresponding trained semantic segmentation algorithm is already integrated in an end-to-end system for MR \cite{gonzalez2022bringing}, making it possible for the user to see her/his own body\footnote{\url{https://youtu.be/XiMmD1UzDiI} please see this video showing the algorithm integrated in MR.}. This is expected to be practically relevant for different MR-enabled applications related to industrial training, education, hybrid conferences or social communication.

The rest of this article is structured as follows: Section~\ref{related_works} describes related works, first concerning different real-time semantic segmentation algorithms. Later Section~\ref{datasets} provides details of the datasets composing EgoBodies. Section~\ref{semantic-segmentation} presents the algorithm considered to segment egocentric bodies in real-time. Then, \ref{results} reports the experimental protocol, segmentation results and the comparison with former segmentation approaches used for MR. Finally, Section~\ref{conclu} concludes the paper with some discussions and future research lines.

\section{Related works}
\label{related_works}

Semantic segmentation architectures are composed of two main subcomponents: an encoder, which is in charge of progressively reducing the spatial information while retaining class information, and a decoder, a component to transform the spatial map from the output of the encoding to the original size of the image, with each pixel containing class information. Long \textit{et al.} \cite{long2015fully} were the first one to propose a semantic segmentation approach based on deep learning. Concretely, they proposed fully convolutional networks (FCNs): a modification of CNN architectures that achieved state-of-the-art performance on semantic segmentation problems using deep learning for the first time. Specifically, they replaced the last fully connected layers of a VGG-16 backbone with fully convolutional ones to preserve the spatial dimension while maintaining class identity information. The decoding subnetwork, placed after the fully convolutional layers, was composed of several upsampling layers to recover original input size. Later on, Badrinarayanan  \cite{badrinarayanan2017segnet} \textit{et al.} introduced the use of a decoder subnetwork similar in number of layers and structure to the encoding layer. Besides, the decoder subnetwork used pooling indices computed in the max-pooling step of the corresponding encoder to perform non-linear upsampling, making it easier for the network to learn how to upsample. 

Later, Zhao \textit{et al.} \cite{zhao2017pyramid} put forward Pyramid Pooling module (PPM) between the encoder and the decoder submodule. This module first computes max pooling to the output of the encoding at different factors, and then concatenates and pass them to the decoding subnetwork. This enables to extract features at different scales, retaining thus both
local and global context. Following the same idea, Chen \cite{chen2018encoder} \textit{et al.}  proposed DeepLabv3+, which made use of Atrous Spatial Pyramid Pooling (ASPP) replacing the former PPM by using dilated convolutions instead of standard ones. Concerning performance, state-of-the-art solutions follows encoder-decoder architecture exploiting object contextual representation through  transformers modules \cite{yan2022lawin}.

\begin{figure*}[t]
  \centering
  \includegraphics[width=0.7\linewidth]{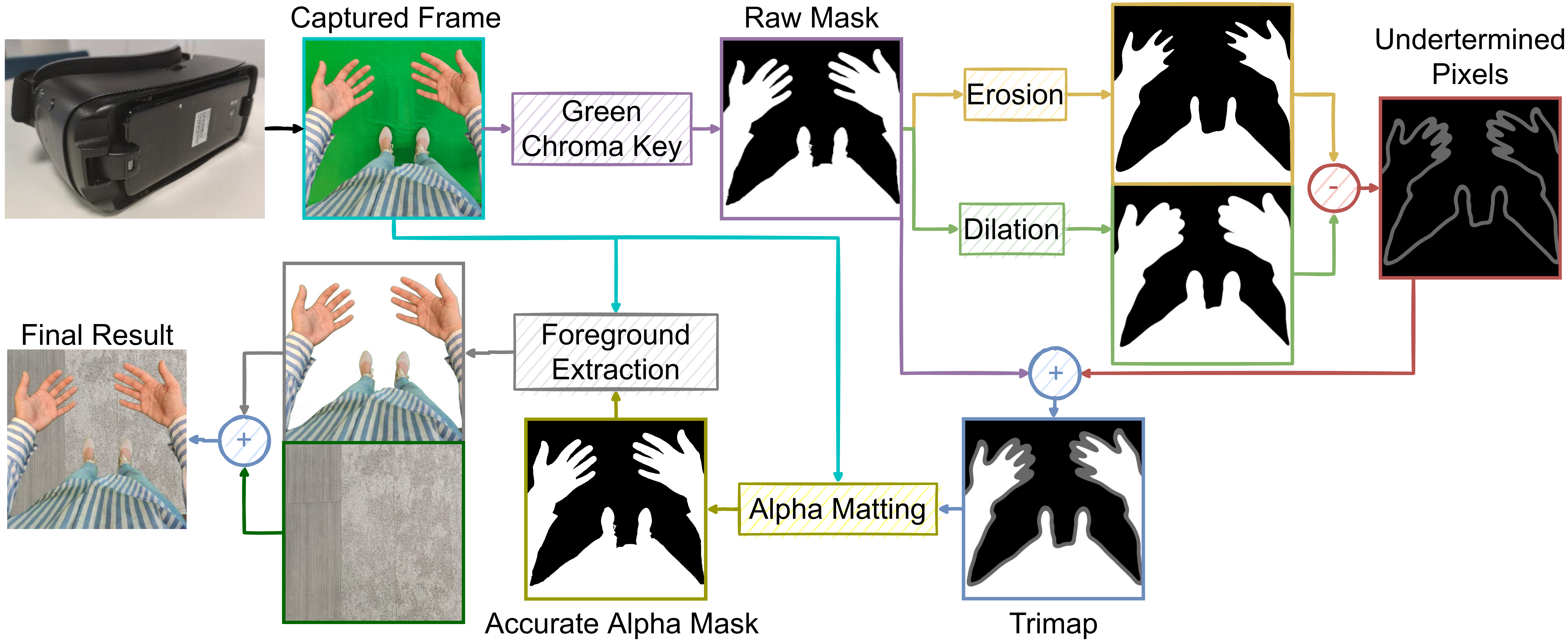}
  \caption{Procedure to obtain semi-synthetic images with automatic groundtruth.}
	\label{fig:ego_human}
\end{figure*}

\begin{figure*}[th]
  \centering
  \includegraphics[width=1.0\linewidth]{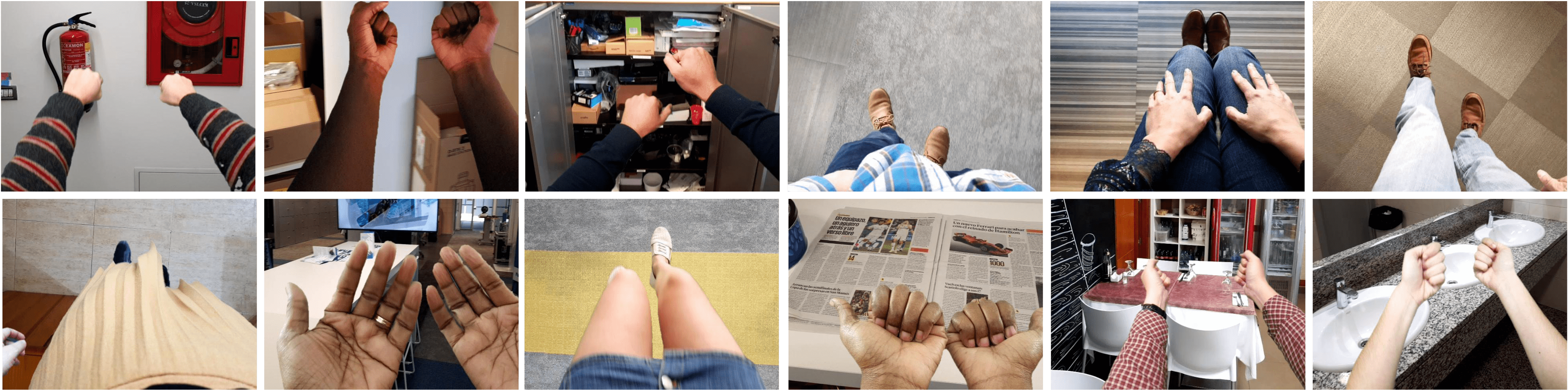}
  \caption{Samples of the semi-synthetic Ego Human images.}
	\label{fig:ego_human}
\end{figure*}

While most semantic segmentation research focuses on improving the accuracy, some attention is also given to find computationally efficient solutions for mobile, real-time, or battery-powered applications. Increasing the network size with layers from the encoding layers increases segmentation accuracy but also increase inference time. Research has been boosted motivated by critical semantic segmentation applications such as autonomous driving. One of the common approaches is to replace deeper encoding networks such as VGG-16 for shallower ones such as Resnet. ENET \cite{paszke2016enet} proposed the use of a few bottleneck modules with $i)$ a single main branch with either the input or a max pooling operation and $ii$) a secondary branch with convolution, batch normalization and Rectified Linear Unit (ReLU). Both branches are later merged back via concatenation. However, performing convolutions in bottleneck modules consume much more runtime memory and thus take longer time, due to the largely expanded number of operations. Later, Zhao et al Zhao \textit{et al.} proposed ICnet architecture \cite{zhao2018icnet}. Rather than using a PPM or APPS, ICNet proposed three encoding networks that extract features at different scales, which are then later fused using a Cascade Feature Fusion module followed by a standard decoder. Later on, Xiang \textit{et al.} proposed ThunderNet architecture \cite{xiang2019thundernet}, which turned out to outperform previous real time semantic segmentation architecture such as ENET \cite{zhao2018icnet} or ICNet \cite{zhao2018icnet}. This architecture is mainly based on three parts: $1)$ a very shallow encoding subnetwork based on the first three Resnet-18 blocks; $2)$ a PPM module after the encoder; and $3)$ a decoding subnetwork. The key benefit of Thundernet is the use of standard convolution layers as that allows to fully optimize adds and multiplications operations when using desktop GPU, as opposed to those networks using bottlenecks.

\section{Egocentric Bodies Dataset}
\label{datasets}

The specific problem of segmenting bodies from an egocentric point of view is quite novel. As a result, none of the benchmarking datasets for semantic segmentation (e.g., PASCAL VOC, Cityscapes) are suitable as training data. Even more related datasets such as EPIC KITCHEN, FPHA, originally created for action recognition, are not suitable since their groundtruth is related to action labels. Due to the lack of proper egocentric human datasets with pixel-wise labelling, and extending previous works  \cite{gonzalez2020enhanced,gonzalez2022real} we decided to create a dataset reflecting real-life settings entitled Egocentric Bodies, composed of three different datasets: EgoHuman \cite{gonzalez2020enhanced}, a subset of THU-READ images \cite{tang2018multi}, and EgoOffices. 

\begin{figure*}[t]
  \centering
  \includegraphics[width=1.0\linewidth]{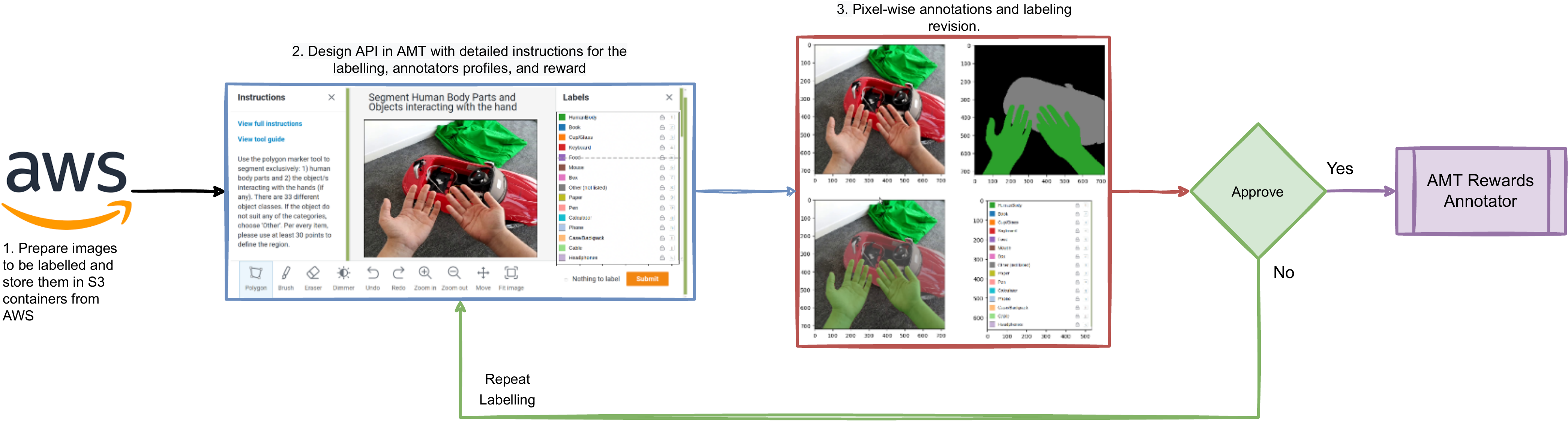}
  \caption{Procedure to get labelling using Amazon Mechanical Turk (AMT) services.}
	\label{fig:amt}
\end{figure*}

\begin{figure*}[th]
  \centering
  \includegraphics[width=1.0\linewidth]{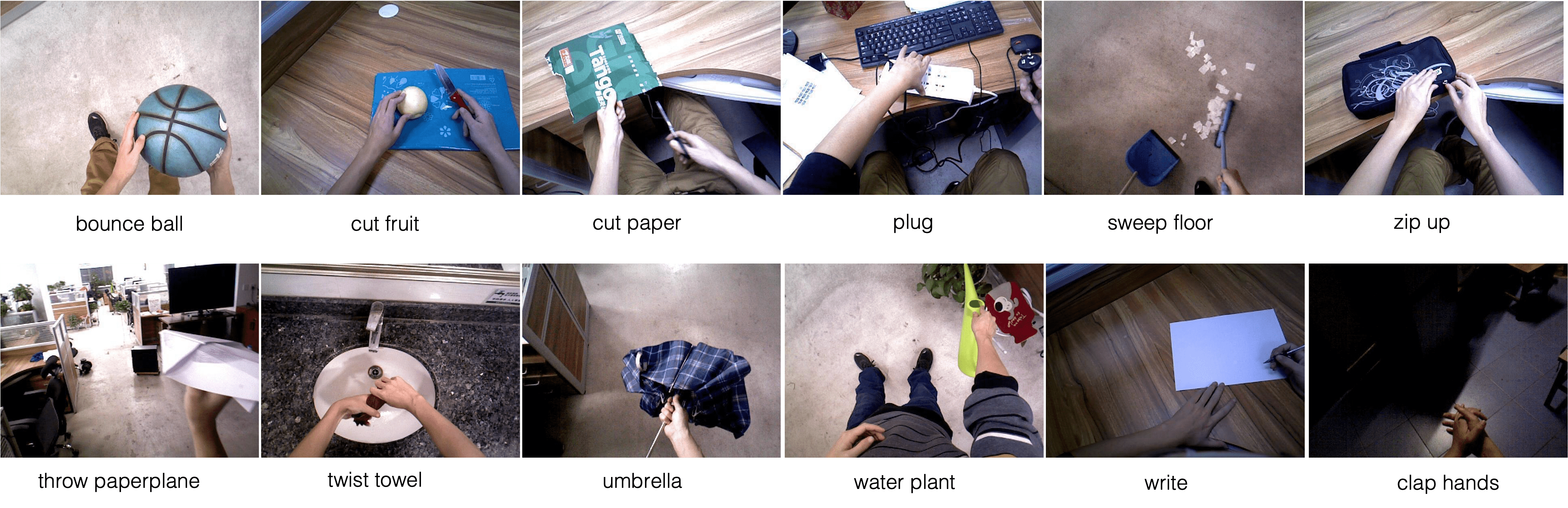}
  \caption{Samples of THU-READ images and their corresponding actions.}
	\label{fig:ego_thuread}
\end{figure*}

\subsection{Ego Human}

Ego Human is a semi-synthetic dataset, which was created with the purpose of easing the labelling process. The specific details on how Ego Human data set was created can be seen in Fig.~\ref{fig:ego_human} and are the following: 
\begin{itemize}
    \item \textbf{Foreground Capturing}: for capturing lower limbs, we asked a total of $13$ users to walk freely through the chroma-key backdrop while being recorded. A second round of data capturing was performed with the users sitting down in a chair also covered by the chroma-key. The recording was done using an Android app installed in a Samsung S8 smartphone placed in the Samsung Gear Framework headset, taking $720\times720$ images at $30fps$. Users repeated the experiments with different outfits including short and long sleeves to add enough variability to the dataset. A total of $6733$ frames from lower limbs were extracted by video sampling. Then, a subset of egocentric arms foreground already available from previous work was added \cite{gonzalez2020enhanced} ($8668$ out of $17233$).  As a result, we have a total of $15,401$ images conforming the Ego Human Segmentation dataset.
    \item \textbf{Background Capturing}: in a second stage, videos of realistic backgrounds were acquired using the same app in three different positions: stand up position looking to the front; stand up position looking to the floor, and sit-down position looking to the floor. A total of $73$, $27$ and $18$ different background videos were acquired, encompassing different indoors scenarios including offices, houses, restaurants, and halls. $2$ frames were sampled per each video. Frames pertaining to the videos looking to the floor were augmented using rotation of $45^{\circ}$, $90^{\circ}$ and $180^{\circ}$.
    \item \textbf{Foreground extraction:} using chroma-key filtering in the HSV color space, we extracted the foreground pixels, corresponding to the human body parts, from the green background as described in \cite{gonzalez2020enhanced}. The result is a mask in which white pixels correspond to the human body parts while the black ones correspond to the background.
    \item  \textbf{Foreground-background realistic blend:} the final step of the Ego Human dataset acquisition is the realistic blend of the foreground frames extracted form the chroma-key background with the backgrounds captured in the posterior step. This blend must be as smooth and realistic as possible for the network to accurately segment unseen real data. To achieve this goal, we use an alpha matting algorithm to accurately estimate the alpha channel values around the foreground edges. More precisely, we used the Shared Sampling Alpha Matting algorithm \cite{gastal2010shared}. For a more detailed description on how to use Alpha Matting, please refer to Appendix A. Fig.\ref{fig:ego_human} depict examples of the resulting semi-synthetic images. 
    \begin{enumerate}
         \item \textit{Trimap image estimation}: the trimap image is a 3-color mask in which the white pixels corresponds to the foreground, black pixel are the backgournd and gray pixels correspond to undetermined pixels. The gray are correspond to the areas around the foreground edge pixels. We estimate the trimap by applying a dilation and erosion kernel on the original mask and subtracting both results, as shown in Fig. \ref{fig:ego_human}. The result of this subtraction corresponds to the gray pixel of the trimap image. 
         \item \textit{Precise foreground alpha channel estimation}: the foreground's alpha channel is estimated using the alpha matting algorithm presented in \cite{gastal2010shared}. The input to this algorithm is the trimap mask and the original captured frame, with the green chroma-key background. The result is a gray-scale mask corresponding to the background alpha channel values, very precise around the edges. The pixel values of the alpha channel mask is normalized between 1 and 0. 
        \item \textit{Foreground-background blending}: in this final step the selected background is realistically blended with the foreground containing the human body parts.Being $\alpha$ the alpha channel mask obtained in the previous step, $\beta$ the background image and $\phi$ the original captured foreground frame, with the green chroma-key background, we estimate the blended final image $\lambda$ as: 
         \begin{equation}
               \lambda_{i,j} = \alpha_{i,j}\cdot\phi_{i,j} + (1-\alpha_{i,j})\cdot\beta_{i,j}
           \end{equation}
        \end{enumerate}
\end{itemize}

\begin{figure*}[th]
  \centering
  \includegraphics[width=0.7\linewidth]{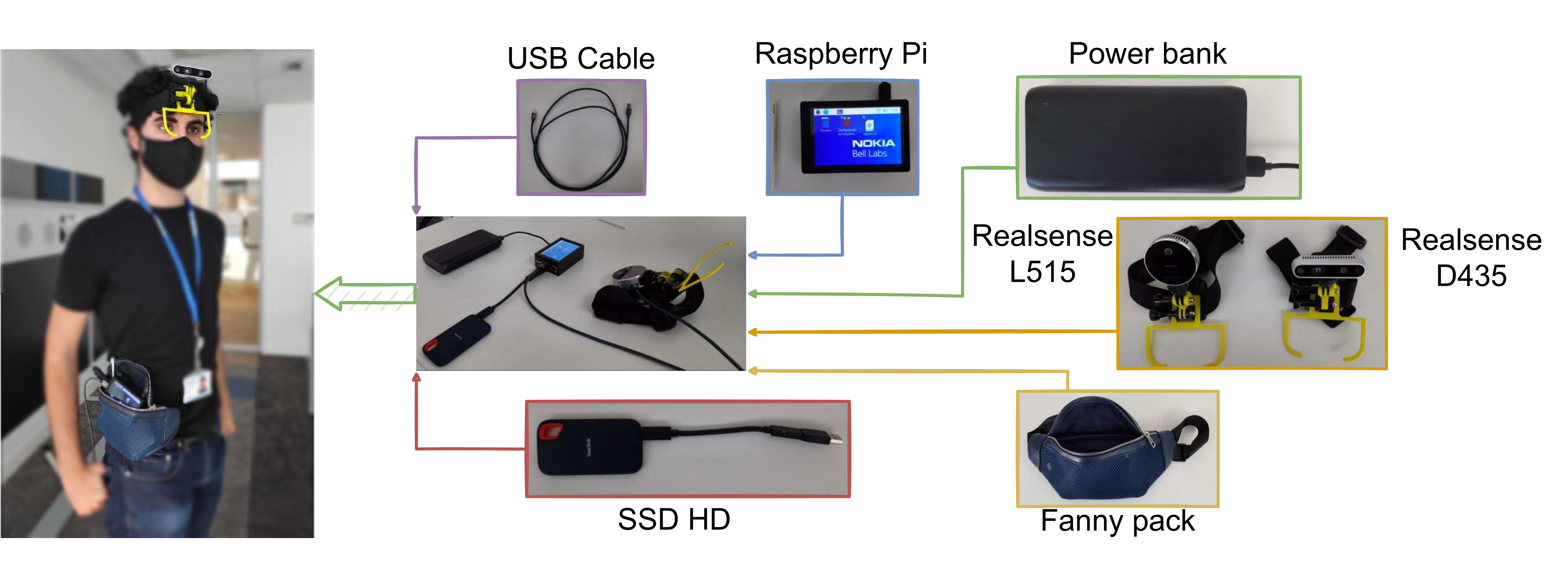}
  \caption{Detailed of the elements conforming the Egocentric Capture Kit for Semantic Segmentation.}
	\label{fig:capture}
\end{figure*}

\begin{figure*}[th]
  \centering
  \includegraphics[width=1.0\linewidth]{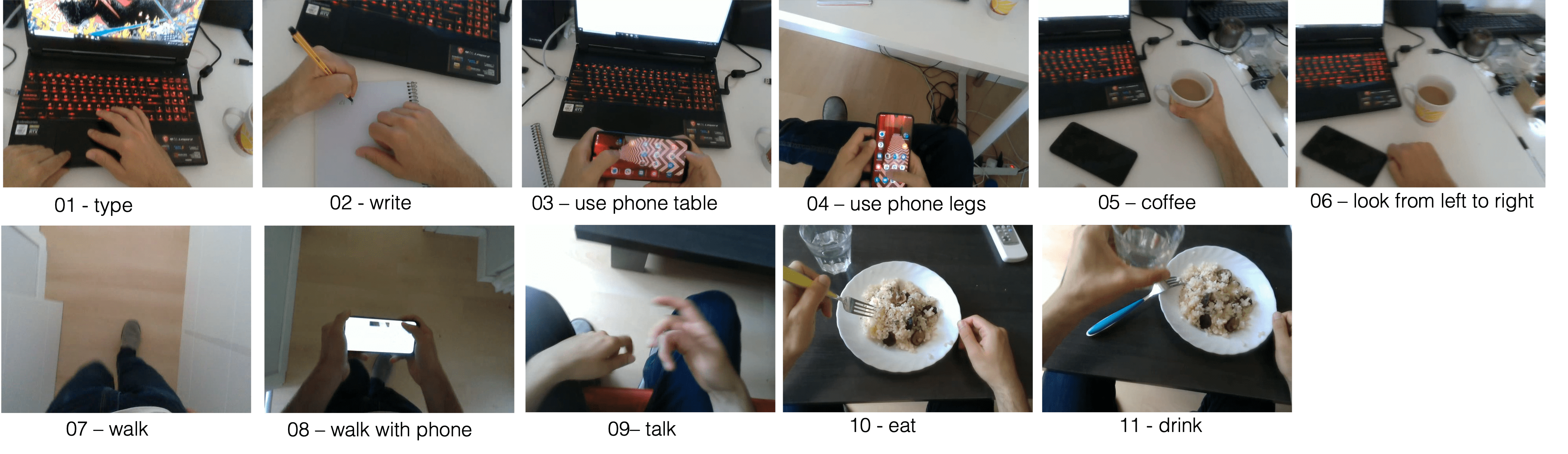}
  \caption{Samples of EgoOffices images and their corresponding actions.}
	\label{fig:ego_offi}
\end{figure*}

\subsection{THU-READ dataset}
\label{thu-read-amt}

THU-READ is a RGB-D dataset collected at Tsinghua University, designed for recognizing egocentric actions which have some relationship with hands \cite{tang2018multi}. It contains recordings of $40$ different actions from $8$ different subjects, repeated $3$ times, making a total of $960$ RGB-D videos.   In this work a representative subset of $640\times480$ images from all users, actions and repetitions is created through video sampling, resulting in a set of $1850$ frames. We designed a labeling tool selected from the semantic segmentation template of Amazon Mechanical Turk (AMT\footnote{https://www.mturk.com/}). In our case, the Human Intelligence Task (HIT) consisted on using the polygon marker tool to define the boundaries of $1)$ human body parts and $2)$ objects interacting with the user. Although not the scope of this work, we also asked Turks to label objects interacting with them into one of $30$ predefined categories. More details on specific objects can be found in \cite{gonzalez2022real}. For details regarding the actions and example images please refer to Appendix A. Fig. \ref{fig:ego_thuread} shows example images performing some of the $40$ possible actions.

From this AMT-based labeling experiment we found of special relevance that: $i)$ better results were obtained from Turks who hold Master Qualification; and $ii)$ a post-processing check of the labels, is required to assure precise boundaries and correct classes\footnote{Done by overlapping labeled images created by Turks on top of their original RGB counterpart images}. If images were not correctly processed, instructors rejected the tasks, providing detailed feedback. Once labeled images were accepted, Turks received a compensation in the range of $25-40$ cents per HIT, comprising also Amazon Fee and 5\% extra for Master Qualification.

\subsection{EgoOffices}

We decided to also include a dataset with real egocentric captures to increase dataset variability in terms of number of users, gender, skin color, scene, objects, and illumination conditions. Data capturing took place between May and June 2021. As can be seen in Fig.~\ref{fig:capture}, the egocentric kit capture was conformed of a raspberry Pi, in charge of managing the logic of capturing egocentric videos from the cameras and handling the writing process in a hard drive with fast read and write speed values. As the idea was to capture egocentric videos of people performing different actions, the egocentric kit required to be portable, therefore, we decided to place all the elements in a belt pouch attached to the user's waist. The recording session was composed of $11$ actions (see Appendix C). To maximize the possibilities of this dataset, we decided to record also both IMU sensors and depth information with 2 different depth sensors: Realsense S435, which estimates depth information through disparity, and Realsense L515, which estimates depth information through Laser Imaging Detection and Ranging (LIDAR) technology. he recording session was composed of the following $11$ actions: $1)$ type in the computer (seated), $2)$ write in a notebook (seated), $3$ use the mobile phone in front of the computer (seated), $4$ use the mobile phone away from the table (seated), $5)$ have a little bit of coffee (seated), $6)$ walk while watching downwards (standing up), $8$ walk while using the mobilephone (standing up), $9)$ sit down in the sofa and chat as if there were someone talking with you, $10)$ eat (seated), $11)$ drink (seated) as can be seen in Fig.\ref{fig:ego_offi}. The recording session took place in the home of the different users and took approximately $40$ minutes. Once captured, we extracted frames from all the different videos of all sensors ($15$-$20$ frames per video), resulting in a total of $8873$ images, coming from $26$ different users ($6$ of them only for SR D435). The associated groundtruth was obtained using AMT services, as reported in Section \ref{thu-read-amt}.

\section{Semantic Segmentation}
\label{semantic-segmentation}

\begin{figure*}[t]
  \centering
  \includegraphics[width=0.7\linewidth]{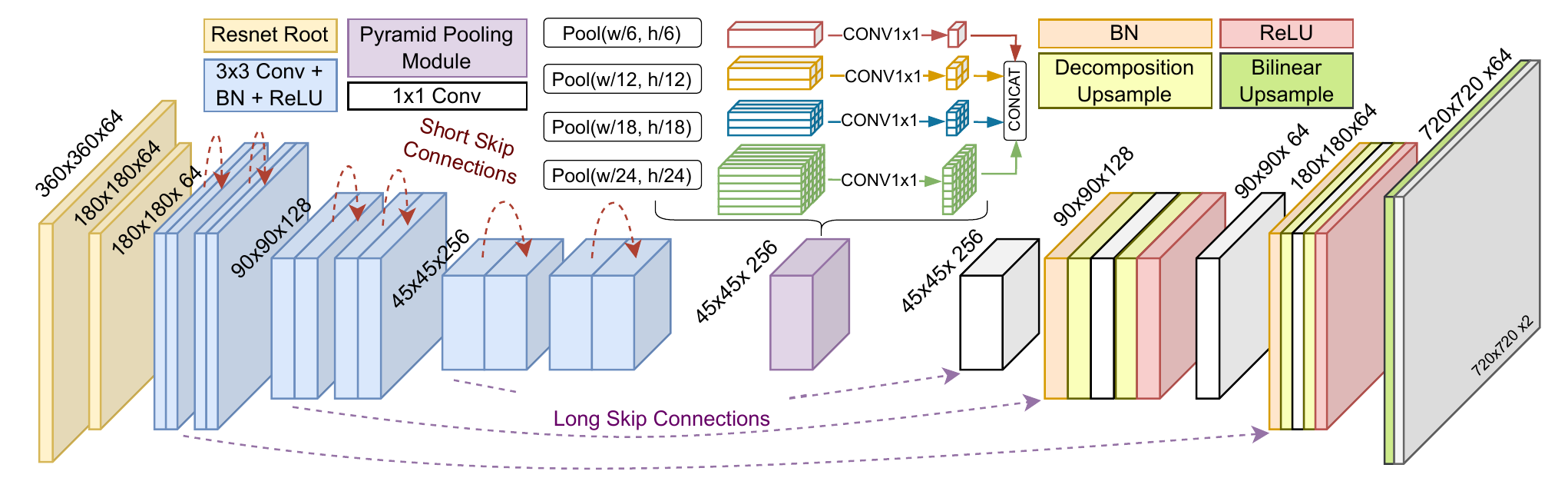}
  \caption{Details of the Semantic Segmentation architecture}
	\label{fig:architecture}
\end{figure*}

Fig.~\ref{fig:architecture} depicts the general architecture designed to segment among two classes: background and human egocentric body parts, which is based on Thundernet architecture \cite{xiang2019thundernet}. Unlike the original network \cite{xiang2019thundernet} and due to the larger size of training images, we decided to use larger sampling pooling factors: $6,12,18,24$. The decoding subnetwork is similar to the one proposed in the original architecture, made up of two deconvolutional blocks. Besides, apart from the skip connections included within the encoding and decoding blocks, we include three more long skip connections between encoding and decoding subnetworks for refining object boundaries. 

This new Thundernet architecture has been developed and trained using Keras framework $2.2.4$ and tensorflow $1.14$. All experiments run on a workstation with 2 NVIDIA GPU 1080 under CUDA $10.0$. The weights from the three Resnet-18 blocks inside encoder are inherited from a model pre-trained on ImageNet dataset. Afterwards, the whole architecture is fine-tuned in an end-to-end approach. Chromatic and cropping augmentation techniques were also applied to the training images. The loss function used was the weighted cross entropy, whose weights were estimated according to the whole frequency of foreground and background pixels in the training set ($0.56$ and $3.27$ for the background and human class, respectively). 

As segmenting egocentric bodies is a $2$-class semantic segmentation problem, we decided to set to $0$ (related to background) all objects from THU-READ and EgoOffices dataset\footnote{This object information will be probably used for future work}.  

\section{Results}
\label{results}

\subsection{Dataset Ablation Studies}
\label{sec:dataset_ablation}
We first trained ThunderNet architecture exclusively with each of the individual datasets conforming EgoBodies. We created training and validation subsets as follows: $12658$ training images and $2743$ validation images for EgoHuman; $1574$ training images (belonging to $7$ of the $8$ users), and $276$ validation images (remainder user) for THU-READ; for EgoOffices, $8078$ images were used for training and $795$ for validation. After many extensive experiments, the hyper-parameters found for the best performance were obtained using an Adam optimizer, a batch size of $4$ (due to the high size of the training images), learning rate of $1e-4$, and weight decay of $2e-4$ for the models trained with the three datasets.

Table~\ref{tab:iou} reports results in terms of Intersection over Union (IoU, see Eq. 1) on the same datasets used in \cite{gonzalez2020enhanced}: GTEA, EDSH, EgoHands and a subset of EgoGesture. As grountruth of the available test datasets presented is only related to hands or skin, but not clothes, reported IoU is underestimated. This means that clothes from arms and torso and lower limbs, even when segmented by our proposed method, count as false positive (FP) and thus, reduce the IoU.

\begin{equation}
    mIoU=\frac{1}{k}\sum_{i=1}^{k}IoU_{i}=\frac{1}{k}\sum_{i=1}^{k}\left[\frac{\mathit{TP}}{\mathit{TP}+\mathit{FP}+\mathit{FN}}\right]_{i}
\end{equation}

As can be seen from Table~\ref{tab:iou}, EgoHuman generalization capabilities are lower (average of $0.32$ of IoU) than the models trained with THU-READ and EgoOffices ($0.47$ and $0.45$, respectively). This might be partially due to the limit number of background images, or the partial realism that semi-synthetic images, which fails to represent reliably real-word settings. It is also worth noting that THU-READ average performance is $15$ percentage points greater than the average performance achieved with EgoHuman, despite containing $10$ times less images than EgoHuman.

\subsection{Results with EgoBodies}

We decided to create EgoBodies as a balanced combination from all three datasets in a $9074$ images dataset ($8005$ for training, $1069$ for validation), considering the performances reported in section \ref{sec:dataset_ablation}. The entire set of $1850$ THU-READ images was included. As for EgoOffices, we reduced the number from $8873$ to $5108$ by keeping $5-10$ frames per video, so that images from EgoOffices and THU-READ were a little bit more balanced in number. We further included a subset of $2116$ specific images from EgoHuman images to increase variability through black ethnicity users, and images from lower limb parts. Same hyperparameters as with the individual datasets were used. As a result, an improvement from $0.47$\footnote{considering the best individual result}, to $0.59$ was achieved, which confirm us the importance of the variability and quality of the datasets. Indeed, gathering all individual datasets, EgoBodies contain images from $47$ subjects ($4$ of them with black ethnicity), each with their representative scene, $51$ different actions, and $4$ different sensors. We also observed that there is still room for improvement in terms of performance achieved with Thundernet, DeepLabv3+ architecture trained with EgoBodies \cite{gonzalez2020enhanced} yields average results of $0.69$ IoU, which outperforms average results from Thundernet with $10$ percentage points. This might be due to its deeper architecture, both at encoding and decoding and with more sophisticated modules such as ASPP, as already described in Section \ref{related_works}. 

Table~\ref{tab:time} indicates inferences times achieved depending on the input resolution. Indeed, inference time of DeepLabv3+ is between $2$ or $3$ times slower than Thundernet. Using Thundernet as segmentation algorithm for Webcam resolution offers $66$ fps, satisfying requirements for MR goggles \footnote{https://developer.oculus.com/resources/oculus-device-specs/}, as opposed to the $23$ fps from DeepLabv3+. 

\subsection{Qualitative Results in the wild}

\begin{figure*}[t]
  \centering
  \includegraphics[width=1.0\linewidth]{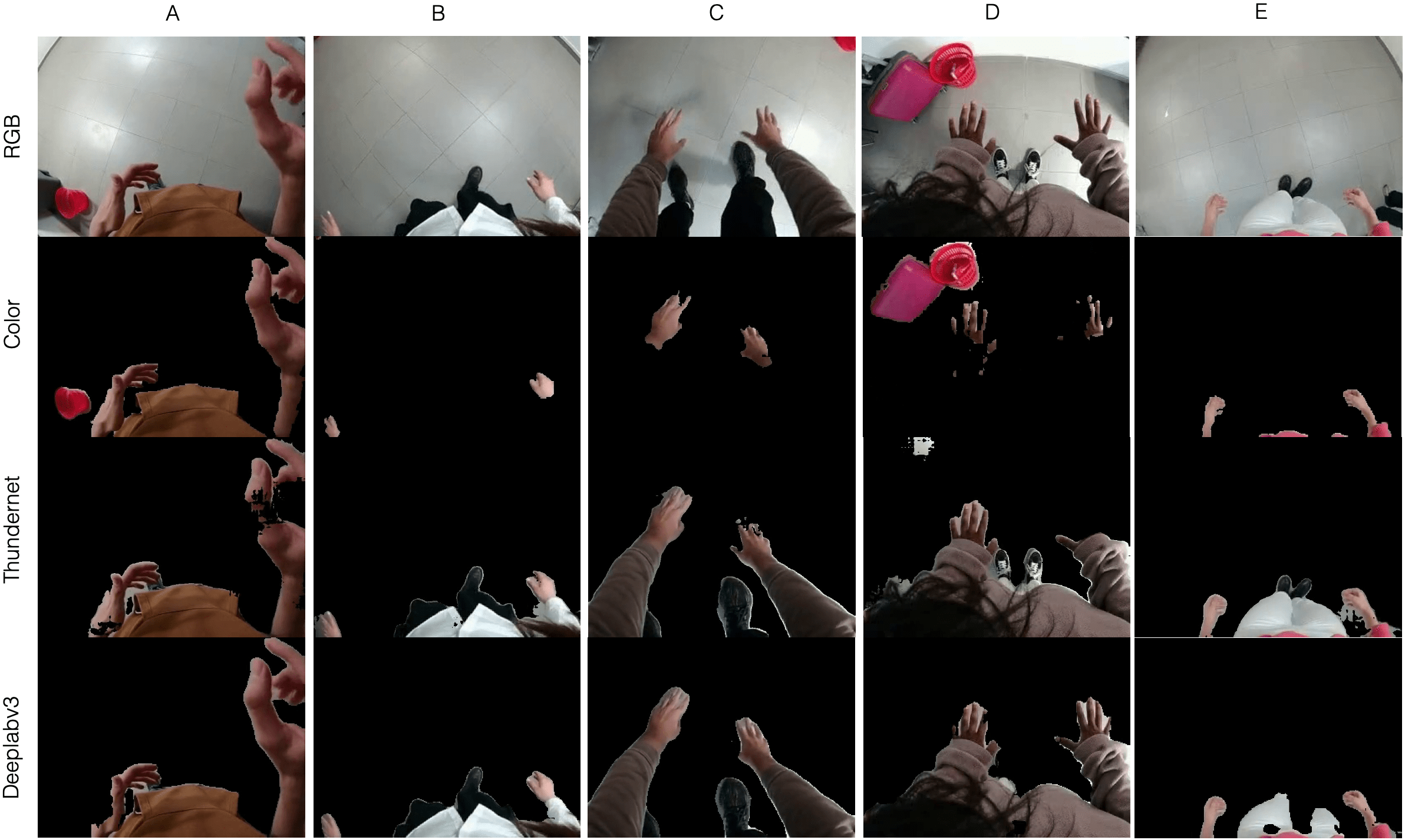}
  \caption{Qualitative results from real egocentric frames and three different segmentation methods: color information, Thundernet, and DeepLabv3+}
	\label{fig:results_quali}
\end{figure*}

As the employed test datasets do not represent reliably the use case of egocentric body segmentation, we captured several egocentric videos from users walking while wearing VR goggles with a stereo camera in front of them, as can be seen in Fig~.\ref{fig:mr}. Fig.~\ref{fig:results_quali} represents several individual frames from those videos\footnote{corresponding videos are included in the supplementary material}, and their corresponding segmentation output, depending on the segmentation algorithm use: skin-color segmentation, Thundernet and DeepLabv3+, both trained with EgoBodies. In general, using color information downgrades severely the results, as only those bodies with favourable clothes will appear (Fig.~\ref{fig:results_quali} A or Fig.~\ref{fig:results_quali} E). Notice also, that both deep learning network manages to segment people regardless of their ethnicity (see for instance Fig.~\ref{fig:results_quali} B and Fig.~\ref{fig:results_quali} D). Likewise, any other item from the scene that will share the key color will be a false positive (segmentation error). Regarding both Thundernet and DeepLabv3+ networks, in general, both networks attempt to segment reasonably well egocentric bodies, although DeepLabv3+ provides a little bit more of precision and less false positives. However, Thundernet is the only solution that satisfy both real time and good quality segmentation. 

\begin{table*}[tb]
\tiny
\centering
  \caption{Results reported in terms of Intersection over Union, including ablation studies to understand the generalization capabilities of each of the dataset conforming EgoBodies and comparison with previous work \cite{gonzalez2020enhanced}.}
  \label{tab:iou}
  \begin{tabular}{cccccc}
    \toprule
        \textbf{Test Dataset} &   \begin{tabular}[c]{@{}c@{}}  \textbf{Thundernet} \\  \textbf{EgoHuman} \end{tabular}   & \begin{tabular}[c]{@{}c@{}}  \textbf{Thundernet} \\  \textbf{THU-READ} \end{tabular}  &  \begin{tabular}[c]{@{}c@{}}  \textbf{Thundernet} \\  \textbf{EgoOffices} \end{tabular}  & \begin{tabular}[c]{@{}c@{}}  \textbf{Thundernet} \\  \textbf{EgoBodies} \end{tabular} & \begin{tabular}[c]{@{}c@{}}  \textbf{DeepLabv3+} \\  \textbf{EgoBodies} \end{tabular}    \\
    \hline
      GTEA &  $0.44$  &  $0.42$ & $0.24$ &  $0.66$ & $0.78$ \\[0.3cm] 
      EDSH2  &  $0.22$  &  $0.55$  & $0.54$ &  $0.65$ & $0.83$ \\[0.3cm] 
      EDSHK  &  $0.22$  &  $0.47$  & $0.46$ &  $0.56$ & $0.64$\\[0.3cm] 
      Ego Hands  &  $0.17$  &  $0.32$  & $0.38$ &  $0.38$ & $0.48$ \\[0.3cm] 
      Ego Gesture  &  $0.25$  &  $0.60$  & $0.66$ &  $0.72$ & $0.76$  \\[0.3cm] 
      Average & $0.32$ & $0.47$ & $0.45$ & $0.59$ & $0.69$ \\
  \bottomrule
\end{tabular}
\end{table*}

\begin{table*}[tb]
\tiny
\centering
  \caption{Inference Times depending on the resolution of the input image. Results are reported on a Xeon ES-2620 V4 @ $2.1$Ghz with $32$ GB powered with $2$ GPU GTX-1080 Ti with $12GB$ RAM.}
  \label{tab:time}
  \begin{tabular}{cccccc}
    \toprule
        \textbf{Resolution} &  Network & 320x240 & 640 x 480 & 960 x 1280 & 1920 x 2560   \\
    \hline
    Inference Time & Thundernet & $7$ ms & $15$ ms & $55$ ms & $254$ ms \\
    Inference Time & DeepLabv3+ & $27$ ms & $42$ ms  & $120$  ms & $460$ ms \\
  \bottomrule
\end{tabular}
\end{table*}

\section{Conclusions}
\label{conclu} 

In this paper, we contributed with a real-time semantic segmentation network that achieves high quality segmentation beyond benchmarking datasets. This is crucial, since the ultimate goal of the algorithm is to be integrated in a MR application where the user wearing the VR goggles can see his/her own body while being immersive in a MR experience. To this aim, we created what we called EgoBodies, an egocentric bodies dataset composed of real egocentric images of different users performing different actions, and a subset of semi-synthetic images to further increase the variability of the dataset. In the future, we plan to extend this work by not only segmenting user's own body, but also objects with which the user interacts with. It would be a real challenge to develop a model that manages to segment objects beyond the ones included in the trained dataset. Probably, one-shot learning or few-shot learning techniques will be of special relevance.

\section*{Appendix A}

THU-READ videos refer to one of the following $40$ egocentric actions: bounce ball, clap hand, close drawer, knock door, lift weights, open door, sweep floor
throw paperplane, thumb, tie shoelaces, twist towel, umbrella, use mobile, water plan, wave hand, wear watch, zip up, clean table, cut fruit, cut paper, draw paper, fetch water, fold, insert tube, manicure, open drawer, open laptop, plug, push button, read book, squeeze toothpaste, stir, tear paper, use chopstick 
use mouse, use stappler, wash hand, wash fruit, wear glove, write, zip up.

\section*{Compliance with Ethical Standards}

The authors declare that the research was conducted in the absence of any commercial or financial relationships that could be construed as a potential conflict of interest.

\bibliography{egbibsample}
\bibliographystyle{abbrv-doi}

\textbf{Ester Gonzalez-Sosa} works in the Extended Reality Lab in Nokia, where she focuses on computer vision applied to Mixed Reality applications, with a focus on real-time, performance in the wild, and application related to fostering human communications.

\textbf{Andrija Gajić} is a computer vision engineer, currently working at AIM Intelligent Machines, where he is responsible for designing and implementing the perception module used in autonomous excavators. In July 2020, Andrija completed an internship at Nokia Bell Labs Spain, focused on applied computer vision in mixed reality.


\textbf{Diego Gonzalez-Morin} works at Extended Reality Lab in Nokia. He is currently pursuing a Ph.D. focused on the application of ultra-dense networks for the implementation of distributed media rendering .

\textbf{Guillermo Robledo} is a systems engineer who holds a masters' degree in Industrial Engineering from the Polytechnic University of Madrid (UPM) since 2021. From November 2020 to August 2021 he was part of Nokia Extended Reality Lab as an intern. This gave him the opportunity to work alongside renowned professionals on providing more functionality to mixed reality applications.

\textbf{Pablo Pérez} is Lead Scientist at Nokia Extended Reality Lab (Madrid, Spain). He is currently leading the scientific activities of Nokia XR Lab, addressing the end-to-end technological chain of the use of Extended Reality for human communication: networking, system architecture, processing algorithms, quality of experience and human-computer interaction. e is Distinguished Member of Technical Staff title from Nokia.

\textbf{Alvaro Villegas} leads the Extended Reality Lab in Nokia, a research center focused in the application of immersive media (VR, AR, XR) to human communications. He is Distinguished Member of Technical Staff title from Bell Labs. In his former role as Head of Bell Labs in Nokia Spain and now as lead of XR Lab he applies XR, AI/ML and 5G/6G technologies to improve human communications.

\end{document}